\RequirePackage[svgnames]{xcolor}

\documentclass[11pt,letterpaper]{mystyle}

\usepackage[all]{hypcap}
\usepackage[svgnames]{xcolor}
\usepackage[comma,authoryear,compress]{natbib}
\usepackage{hyperref}
\hypersetup{
    colorlinks = true,
    citecolor = {YaleBlue},
    linkcolor = {YaleBlue},
    urlcolor  = {YaleBlue},
}

\usepackage{graphicx}
\usepackage{booktabs}
\usepackage{amsmath}
\usepackage{amssymb}
\usepackage{mathtools}
\usepackage{amsthm}
\usepackage{enumitem}
\usepackage{subcaption}
\usepackage{cleveref}
\usepackage{float}
\usepackage{placeins}   
\usepackage{textcomp}   
\usepackage{xurl}       
\usepackage{stfloats}   
\usepackage{flafter}    
\usepackage{adjustbox}
\usepackage[font=small,labelfont=bf]{caption}

\definecolor{blanchedalmond}{rgb}{1.0, 0.92, 0.8}
\definecolor{carmine}{rgb}{0.59, 0.0, 0.09}
\definecolor{lightblue}{rgb}{0.22,0.45,0.70}%

\renewcommand{\mathbf}{\boldsymbol}

\makeatletter
\def\Ddots{\mathinner{\mkern1mu\raise\p@
\vbox{\kern7\p@\hbox{.}}\mkern2mu
\raise4\p@\hbox{.}\mkern2mu\raise7\p@\hbox{.}\mkern1mu}}
\makeatother

\definecolor{amaranth}{rgb}{0.9, 0.17, 0.31}
\definecolor{antiquebrass}{rgb}{0.8, 0.58, 0.46}
\definecolor{antiquefuchsia}{rgb}{0.57, 0.36, 0.51}
\definecolor{chromeyellow}{rgb}{0.31, 0.47, 0.26}

\title{FromPitch2Board: Benchmarking LLM Agents\\ in Long-Horizon Football Management}

\runningtitle{FromPitch2Board}

\author[1,2]{Peiyu Zang}

\affil[1]{School of Mathematical Sciences, Beijing Normal University}

\affil[2]{CoRe Lab, Institute for Artificial Intelligence, Peking University}

\correspondingauthor{Peiyu Zang, \texttt{202421130105@mail.bnu.edu.cn}}

\begin{document}

\begin{abstract}
Long-horizon agent benchmarks typically report how far an agent progresses,
but do not identify whether its performance comes from the foundation model,
scaffold, responsibility scope, match-control granularity, or horizon. We
introduce FromPitch2Board, a deterministic football-management benchmark that
studies five configurable factors through controlled comparisons on a single
simulator, using paired seeds and a frozen calibration. We evaluate four
foundation models and four agent scaffolds. In the Model Track, Coach points
Z-scores span $0.19$, while Manager points Z-scores span $0.68$, with GPT-5.6
showing a sharp rise in passivity under responsibility expansion. Its
responsibility ladder rises from $46.1$ to $58.1$ points with recruitment, then
falls to $46.8$ under full management, localizing the regression to the final
responsibility boundary. Across that boundary, its skipped-decision rate rises
from 1.1\% to 57.9\%. Within the Flash--Pro pair crossed across every scaffold,
scaffold choice changes Manager points Z-scores by up to $0.48$ relative to the
fixed stateless scaffold. The 3Y cohort shows a directional reversal in mean
ranking between years one and three, while a selected Claude Code+Pro
configuration peaks in year three and remains below that peak, showing that
responsibility scope and horizon expose behavior changes that a single headline
score conceals.

\vspace{2mm}


\vspace{3mm}

\textbf{Code}: \href{https://github.com/factnn/FromPitch2Board}{https://github.com/factnn/FromPitch2Board}

\end{abstract}

\maketitle
\vspace{1mm}

\begin{figure}[!ht]
\centering
\includegraphics[width=0.80\textwidth]{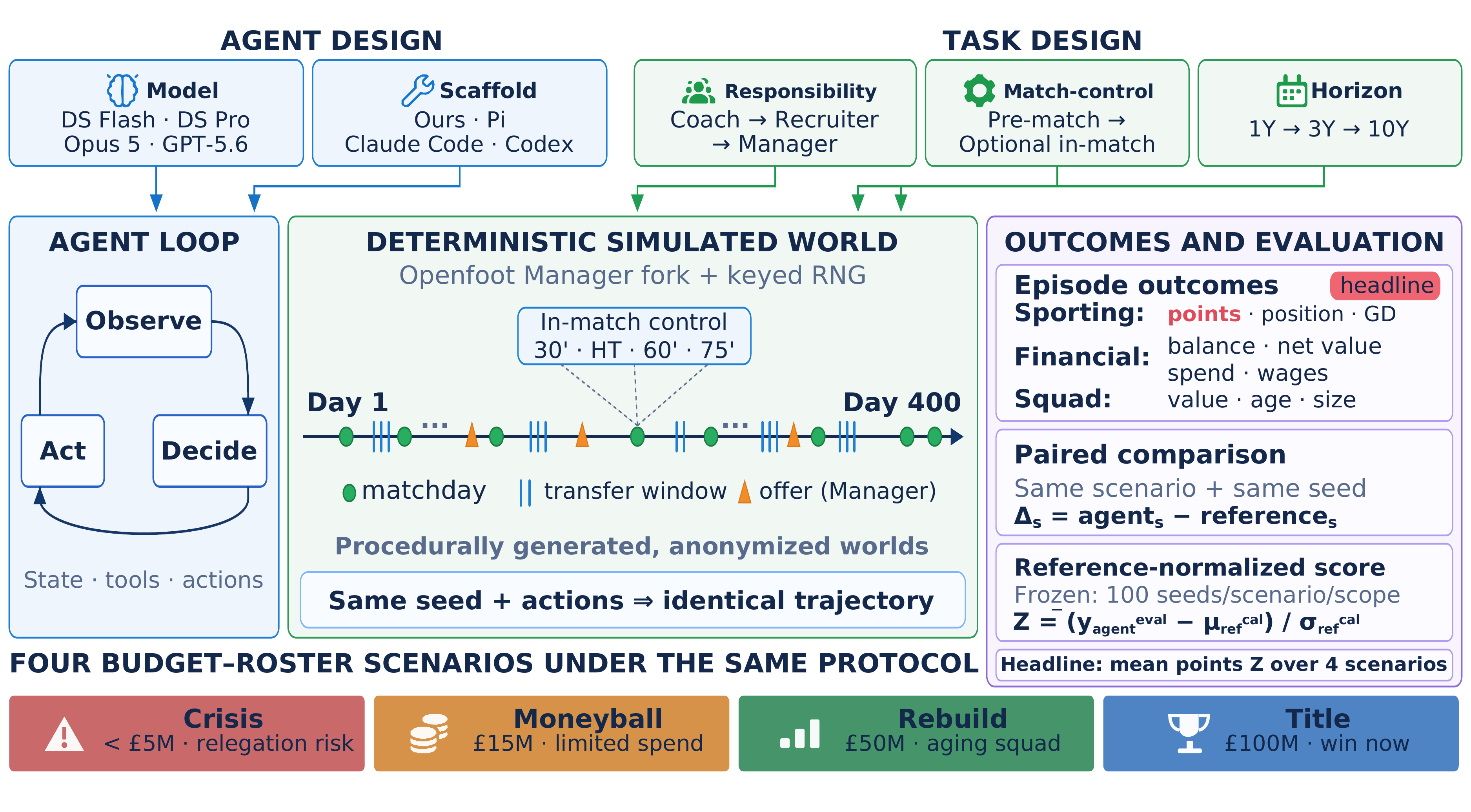}
\caption{\textbf{FromPitch2Board separates five experimental factors.}
Responsibility and horizon operationalize functional composition and temporal
persistence; match-control granularity configures decision density,
while model and scaffold support attribution under paired-seed evaluation.}
\label{fig:overview}
\end{figure}

\vspace{1mm}

\section{Introduction}
\label{sec:intro}

Large language models are increasingly evaluated as agents that interact with
software, people, and persistent worlds~\citep{agentbench,gaia}. Long-horizon
agency poses two distinct problems. \emph{Temporal persistence} asks whether
competence survives recurring, state-dependent decisions whose consequences
accumulate. \emph{Functional composition} asks whether competence on an
existing responsibility set is preserved when additional heterogeneous
responsibilities are assigned to the same agent. WebArena and OSWorld test realistic computer
use~\citep{webarena,osworld}; SWE-bench tests whether models can resolve
repository-scale issues~\citep{swebench}; and $\tau$-bench and ToolSandbox
evaluate stateful tool use and user interaction~\citep{tau,toolsandbox}.
These benchmarks have made agent evaluation substantially more realistic, but
their episodes remain bounded around a specified task. They do not directly
test functional composition and temporal persistence in one controlled setting.

Our results make this distinction concrete. GPT-5.6 records the highest
Model-Track Coach score, although the four models are separated by only $0.19$
Z. Giving it recruitment authority improves
its season points, but adding full-management responsibilities removes that
gain and returns its sporting performance near the greedy reference. Across this last boundary, its skipped-decision rate---the share of decision
points at which it takes no action and continues---rises from 1.1\% to 57.9\%,
while
invalid actions remain at zero. A single-role score therefore conceals both
where performance changes and how the degradation appears in behavior. This paper studies how behavior changes as responsibility expands beyond a
narrower responsibility scope.

\paragraph{Why existing benchmarks do not measure this.}
Several lines of work capture pieces of the problem. Planning benchmarks test
long action sequences~\mbox{\citep{deepplan,planbench}}, while ultra-long-horizon
suites extend them to more persistent settings~\mbox{\citep{lhtb,ultrahorizon}}.
Memory benchmarks instead isolate retention over extended
interactions~\citep{longmemeval,memgym}.
Game environments provide partial observability and sequential consequences,
but generally expose a fixed player role~\citep{grf,balrog,textarena}.
Long-horizon business simulations introduce accumulated resources and delayed
objectives: Vending-Bench studies operational coherence~\citep{vending}, while
CEO-Bench and YC-Bench evaluate organizational decision making over simulated
time~\citep{ceo,yc}. FM-Bench is the closest concurrent work, evaluating
full-control football managers over as many as twenty seasons~\citep{fmbench}.
These designs establish that long-horizon management is difficult, but do not
jointly vary model, scaffold, responsibility scope, match-control granularity,
and horizon within one environment. An endpoint score cannot attribute
performance to a factor.

\paragraph{Our question and design requirements.}
We ask whether competence is preserved as responsibilities expand and persists
across time, and how model and scaffold shape both outcomes. This requires more
than extending episode length. The environment must preserve
partial information, a functioning transfer economy, and competing sporting
and financial objectives. Candidate and reference policies must face matched
exogenous randomness, while normalization must remain stable across scenarios.
Finally, the agent interface must itself be treated as an experimental factor:
SWE-agent shows that agent--computer interfaces affect software-engineering
outcomes~\citep{sweagent}, and Continual Harness studies adaptation around a
fixed foundation model~\citep{continual}. AgentScope and Agent Lightning likewise
make the surrounding agent system an explicit object of design or optimization
~\citep{agentscope,agentlightning}. Pok\'eAgent similarly distinguishes
model and agent tracks~\citep{pokeagent}, motivating a controlled cross rather
than attributing every observed difference to the model alone.

\paragraph{FromPitch2Board.}
We build FromPitch2Board (Figure~\ref{fig:overview}) on Openfoot
Manager~\citep{ofm}, an independently
developed football-management simulator that we make deterministic using
keyed random streams throughout world generation, matches, and turn-level
simulation. Responsibility from Coach through Recruiter to Manager tests
functional composition, while horizons from one to ten seasons test temporal
persistence. A separately configurable match-control axis varies decision
granularity from pre-match choices to live intervention. A Model Track fixes a stateless scaffold while changing the
foundation model; an Agent Track evaluates Ours, Pi, Claude Code, and Codex
under a common interface. Headline comparisons use paired seeds and frozen,
responsibility-scope- and scenario-specific reference distributions
calibrated over 100
seeds. This factor-controlled design exposes responsibility boundaries, scaffold
effects, and temporal degradation that a single endpoint score obscures.

Our contributions are four:
\begin{itemize}
\item \textbf{A deterministic compositional-management benchmark} that varies
model, scaffold, responsibility scope, match-control granularity, and horizon within
one simulation and protocol.
\item \textbf{Functional-composition diagnostics}, including the Composition
Gap and a responsibility ladder that localizes GPT-5.6's sporting regression
to the Recruiter--Manager boundary.
\item \textbf{Temporal-persistence diagnostics} showing a directional reversal
in mean ranking within the three-year cohort and that a ten-year
trajectory peaks early and remains below that peak.
\item \textbf{Controlled model--scaffold attribution within a fully crossed
Flash--Pro pair}.
\end{itemize}

\section{Related Work}
\label{sec:related}

\paragraph{Long-horizon management.}
Business simulations evaluate resource allocation and organizational
decisions over extended horizons~\citep{enterprise,merchant,coffee,stock},
with CEO-Bench and YC-Bench focusing on startup management~\citep{ceo,yc}.
FM-Bench is the closest concurrent comparison: it evaluates full-control
football managers using deterministic replay, scripted and human baselines,
and a privileged-information oracle~\citep{fmbench}. FromPitch2Board asks a
different question. Rather than fixing the management role, it varies
responsibility scope, match-control granularity, and horizon within the same world to
locate where performance changes.

\paragraph{Attributing agent performance.}
Agent outcomes reflect both the foundation model and the surrounding
interface. Harness-Bench and The Scaffold Effect directly cross foundation
models with execution harnesses on coding tasks~\citep{harnessbench,scaffoldeffect}.
SWE-agent and Continual Harness provide earlier evidence for interface design
and adaptation~\citep{sweagent,continual}, while Pok\'eAgent separates model and
agent tracks across game tasks~\citep{pokeagent}. Planning and memory benchmarks
isolate particular cognitive demands~\citep{deepplan,longmemeval,memgpt}, but do
not cross them with management scope. FromPitch2Board combines these
perspectives by crossing model and scaffold while independently configuring
responsibility scope, match-control granularity, and horizon.
See Appendix~\ref{app:benchmark_comparison}.

\section{FromPitch2Board: Task and Benchmark Design}
\label{sec:benchmark}

FromPitch2Board is a single simulation of professional club football management
with three separately configurable task axes: responsibility scope,
match-control granularity, and horizon. Responsibility and horizon test
functional composition and temporal persistence, respectively; match-control
granularity varies the density and timing of decisions.

\subsection{Task axes for composition and persistence}

FromPitch2Board configures responsibility scope, match-control granularity, and
horizon on the same simulated world. The \emph{responsibility
scope} axis is a strict three-rung ladder. \textbf{Coach} selects the starting
eleven, manages rotation and squad condition across matchdays, and chooses
pre-match tactics, with transfers frozen. \textbf{Recruiter} adds the buying
side of the market: scouting under partial observability and bidding, while
selling remains unavailable. \textbf{Manager} adds incoming-offer negotiation,
transfer-listing, and responsibility for wage and transfer budgets. Moving up
this ladder changes the available responsibilities without changing the world,
opponent, horizon, or scoring.

Expanding responsibility can introduce decision
opportunities, such as incoming offers; we treat these as part of the added
management scope rather than as a separate cadence intervention.

The \emph{match-control granularity} axis is separate. Its default setting
allows pre-match lineup and tactical decisions. Enabling in-match control adds
checkpoints at the 30th minute, half-time, the 60th minute, and the 75th minute;
the agent then observes the live score, phase, and player condition and may
substitute players or change formation and play style. Finally, the
\emph{horizon} axis extends an otherwise fixed Manager configuration from one
season to three or ten consecutive seasons, allowing squad aging, contract
expiry, and long-term finances to accumulate.

\subsection{Decision cadence and partial observability}

Time advances day by day and pauses at decision points: every matchday of the
managed club (all responsibility scopes), roughly every three days during open
transfer windows (Recruiter and Manager), and whenever a fresh transfer offer
arrives (Manager). The managed club
is one actor in a living economy rather than a sandboxed object: every other
club is run by its own AI that buys, sells, and adapts within the same
simulated economy. Such an active economy is a property shared by
recent management benchmarks~\citep{fmbench,vending} and a precondition for
our setting: a benchmark whose opponents are static fixtures measures
solitaire, not management. A single season of four
hundred simulated days contains between roughly forty and fifty decision
points; a decade-long horizon multiplies that by ten, and in-match control adds
four stops per matchday. The agent observes its own squad's ratings, while
market-player abilities remain hidden until scouting, which provides noisy
estimates.

\subsection{Scenarios and tracks}

Four budget-and-roster scenarios instantiate the management spectrum:
\textit{crisis} (a relegation-zone club with a transfer budget under
\pounds 5M), \textit{moneyball} (mid-table, \pounds 15M, a board objective of a
top-half finish while keeping net spend under \pounds 5M), \textit{rebuild}
(an aging squad with a \pounds 50M budget and a top-four objective), and
\textit{title} (a strong squad with a \pounds 100M budget, expected to win).
Each scenario fixes club strength rank and budgets, so difficulty is a property of the
world, not of the prompt.

The \textbf{Model Track} fixes our stateless scaffold (observe, then take one
action, with no session memory) and swaps only the foundation model,
isolating the foundation-model effect under a fixed scaffold.
The \textbf{Agent Track} evaluates four named scaffolds: Ours, Pi, Claude
Code, and Codex. Ours is our standardized stateless observe--act scaffold.
The other three are agentic coding CLIs: Pi deliberately uses a minimal agent
design, whereas Claude Code and Codex are full-featured coding agents. All
receive the same task interface and evaluation protocol while retaining their
native tool loops and context management. Every scaffold runs in its own
workspace outside the simulator and cannot read simulator source, internal
state, or other runs. This measures end-to-end scaffold-configuration effects under a shared
external task protocol. The tracks together implement the attribution question of
Section~\ref{sec:intro}.

\subsection{Environment: deterministic, agent-native}

FromPitch2Board builds on Openfoot Manager~\citep{ofm}, an independently
developed, open-source, agent-facing football-management simulator with a
player-level match engine and transfer economy. We retain its text interface
and add keyed random streams throughout world generation, matches, and
turn-level simulation, producing bit-identical replay under a fixed action
sequence. Because the simulator predates this benchmark, its dynamics were not designed for the tested agents. Evaluation worlds, players, and
club identities are procedurally generated and anonymized, reducing direct
instance-level contamination. Experiments use the medium world of three
national leagues (about 120 clubs), with simulator-only seasons taking five
seconds.

\begin{table*}[t]
\centering
\small
\setlength{\tabcolsep}{9pt}
\begin{tabular}{lllrrr}
\toprule
Track & Scaffold & Model & Coach Z & Manager Z & Composition Gap\\
\midrule
Model & Ours & DS Flash & $+0.51$ & $+0.67$ & $-0.16$\\
      & Ours & DS Pro   & $+0.59$ & $+0.76$ & $-0.17$\\
      & Ours & Opus 5   & $+0.55$ & $+0.61$ & $-0.06$\\
      & Ours & GPT-5.6  & $\mathbf{+0.70}$ & $+0.08$ & $\mathbf{+0.62}$\\
      & Rule policy & Greedy reference & $+0.08$ & $-0.09$ & $+0.17$\\
\midrule
Agent & Pi & DS Flash & $+0.55$ & $+1.08$ & $-0.53$\\
      & Pi & DS Pro   & $+0.55$ & $\mathbf{+1.17}$ & $-0.62$\\
      & Claude Code & DS Flash & $+0.54$ & $+1.02$ & $-0.48$\\
      & Claude Code & DS Pro   & $+0.46$ & $+0.95$ & $-0.49$\\
      & Codex & DS Flash & $+1.05$ & $+1.15$ & $-0.10$\\
      & Codex & DS Pro   & $\mathbf{+1.39}$ & $+1.14$ & $+0.25$\\
\bottomrule
\end{tabular}
\caption{\textbf{Main leaderboard across model, scaffold, and responsibility.}
Rows average four scenarios with eight seeds each against the frozen calibration; bold marks per-track extrema.}
\label{tab:main}
\end{table*}

\subsection{Evaluation protocol}

\paragraph{Metrics.} A finished episode records ten raw outcomes: points,
final position, and goal difference; balance, net value, net transfer spend,
and wage bill; and squad value, average age, and squad size. Net value is the
change in club net worth, cash plus squad value, relative to the start of the
episode, so multi-season curves remain anchored to the same initial state. Points is the headline ranking metric; every
ranking here is a points statistic. Other raw outcomes and derived
budget-violation and squad-size diagnostics are auxiliary and are not combined
into an aggregate score. For Manager and Recruiter these cover net value, squad
value, transfer- and wage-budget violations, and squad size against the 22--26
target; for Coach, where transfers are frozen, only average age. We report them
alongside points because sporting, financial, and squad-health outcomes need not
move together.

\paragraph{Paired evaluation and reference-normalized scoring.} Raw scores
across seeds are noisy, so the candidate and reference run the same evaluation
seeds in the same scenario and world. For seed $s$, we report the paired
difference $\Delta_s=y_s^{\mathrm{agent}}-y_s^{\mathrm{ref}}$. For central paired
contrasts, we report $\bar{\Delta}=n^{-1}\sum_s\Delta_s$ with uncertainty
stated explicitly (mean$\pm$SE unless noted). This matching controls variance by aligning seeded exogenous randomness
across policies. Separately, we freeze a one-hundred-seed
calibration distribution for the greedy reference per scenario and
responsibility scope.
The normalized score is
\[
Z = \frac{\bar{y}^{\mathrm{agent}}_{\mathrm{eval}}
        -\mu^{\mathrm{ref}}_{\mathrm{cal}}}
       {\sigma^{\mathrm{ref}}_{\mathrm{cal}}}.
\]
Thus, the paired protocol supports low-variance raw comparisons, while $Z$
places candidates on a stable reference scale. Table scores equally average the
four scenario-specific Z-scores.

\paragraph{Resource-use accounting.} The runner records input, output, and
cache-read tokens, wall-clock time, and invalid-action rate where available.
Costs are converted to a common pricing baseline
(pre-August-17 list prices for DeepSeek and undiscounted official list prices for Claude and GPT).

\paragraph{Determinism.} The simulator uses keyed random streams, yielding
deterministic trajectories under fixed seeds and action sequences; horizons
compose consistently across multi-season runs. The keyed RNG partitions
randomness into semantic sub-streams for days, matches, and scouting, so
consuming randomness in one domain does not shift unrelated draws in another.
The guarantee covers the simulator; hosted-model inference is not part of it and
may vary across repeated calls.

\section{Experiments}
\label{sec:experiments}

\textbf{Setup.} Unless noted otherwise, one-season results use the medium
world, four scenarios, and eight paired seeds (42--49) under the scoring
protocol of Section~\ref{sec:benchmark}. The Model Track compares DS Flash,
DS Pro, Opus~5, and GPT-5.6 with the Ours scaffold; the Agent Track compares
Ours, Pi, Claude Code, and Codex. The Ours+Flash and Ours+Pro cells are shared
between the two tracks, giving ten unique configurations. Each contains 64
episodes (four scenarios, eight seeds, and two responsibility scopes), for 640
episodes in the main leaderboard. Exact model identifiers and reasoning settings appear
in Appendix~\ref{app:protocol}.

The results follow two primary questions: whether competence is preserved as
responsibilities expand and whether it persists across seasons. Model--scaffold
crosses support attribution; match-control, memory, cost, and resource use are
supporting diagnostics.

\begin{figure*}[t]
\centering
\includegraphics[width=\textwidth]{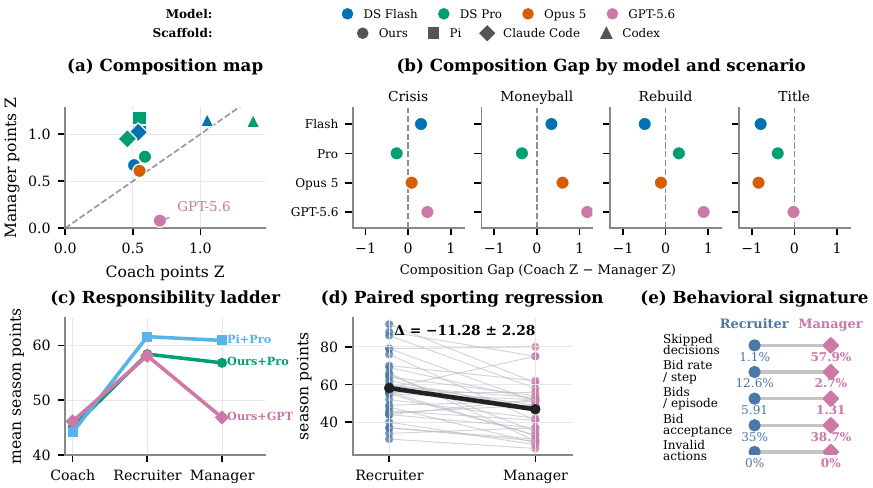}
\caption{\textbf{Responsibility expansion exposes a model-dependent behavioral signature.}
Panels connect the composition map and scenario gaps (a--b) to responsibility
ladders, 32 paired GPT-5.6 episodes, and the resulting behavior (c--e).}
\label{fig:behavior}
\end{figure*}

\subsection{RQ1: Functional composition across responsibilities}
\label{sec:rq1}

The Model-Track rows of Table~\ref{tab:main} report Coach and Manager Z-scores for the four models.
The Coach scores occupy a narrow range: the best and worst models are separated by only $0.19$ Z. The Manager scope is
more dispersed, with a $0.68$-Z range driven by GPT-5.6's score of $+0.08$;
the other three models remain within $0.15$ Z. DS Pro achieves the highest mean
Model-Track Manager score at a lower cost than Opus~5 and GPT-5.6. The composition axis shows behavior that the Coach
ranking hides. We define the \emph{Composition Gap} as
$Z_{\mathrm{Coach}}-Z_{\mathrm{Manager}}$, so a positive value denotes a decrease
in normalized advantage relative to the scope-specific reference after management
responsibilities are added. Accordingly, the GPT-5.6 boundary result is
established by the paired raw Recruiter-to-Manager contrast, and the
Composition Gap serves as a complementary reference-normalized diagnostic. Flash, Pro, and Opus 5 remain
flat or improve relative to their scope-specific references ($-0.17$ to
$-0.06$), whereas GPT-5.6 has a gap of $+0.62$ and falls to the level of the
greedy reference. With Coach and Manager calibrated separately, the gap measures a
within-model change in reference-normalized advantage rather than raw points.

\textbf{Finding 1.} \emph{Under an aligned protocol, Model-Track Coach scores span
$0.19$ Z, but responsibility expansion separates models:
three retain or improve their relative advantage, while GPT-5.6 loses $0.62$
Z. The Composition Gap exposes a change in reference-normalized advantage
that the single-role ranking hides.}

To localize where added responsibilities change behavior, the responsibility ladder evaluates three rungs on the same scaffold: coach-only, coach-plus-recruitment (buying, no
selling), and full manager. Mean season points across the ladder are
$45.0 \to 58.4 \to 56.8$ (Figure~\ref{fig:behavior}): buying is a net gain
of 13.4 points, while the full-management rung costs 1.6 points
but raises net value by \pounds 6.8M and reduces squad
size. Repeating the ladder with the Pi scaffold yields
$44.1 \to 61.6 \to 60.9$ points. Thus, across both tested scaffolds, the
large change occurs when buying is enabled, whereas full management
changes mean points by less than two. We additionally evaluate the middle rung for GPT-5.6. Its raw-point trajectory is $46.1 \to 58.1 \to 46.8$: recruitment
improves performance by 12.0 points, but the gain disappears under the bundle
of full-management responsibilities. In a paired comparison, Recruiter exceeds Manager by
$11.28 \pm 2.28$ points (mean $\pm$ SE, $n=32$). Thus the GPT-5.6 sporting
regression is localized to the Recruiter-to-Manager responsibility boundary, not
to recruitment access itself. Across that boundary, its skipped-decision rate
rises from 1.1\% to 57.9\%, while its bid rate falls from 12.6\% to 2.7\%
(5.91 to 1.31 bids per episode), so participation also declines in absolute
count; both configurations have zero invalid
actions. Among bids with an explicit accepted/rejected result, acceptance does
not fall (35.0\% versus 38.7\%). The sporting regression therefore has a behavioral
signature of passivity rather than malformed or invalid actions.

The regression is sporting rather than universal. Relative to Recruiter,
Manager loses 11.28 points, 4.19 league places, and 21.75 goal-difference
units, while spending \pounds5.73M less and reducing wages by \pounds33.1k.
Cash rises by \pounds5.79M, but net value does not improve
($-\pounds0.83$M $\pm$ \pounds2.07M SE), and squad value falls by
\pounds6.61M. Full management therefore induces financial retrenchment, not a
uniform outcome collapse or a compensating increase in net value.

\textbf{Finding 2.} \emph{Responsibility effects are model-dependent. For
GPT-5.6, recruitment raises season points, while full-management
responsibilities remove the entire gain; its boundary traces exhibit a sharp
rise in passivity rather than invalid actions.}

\begin{figure*}[t]
\centering
\includegraphics[width=\textwidth]{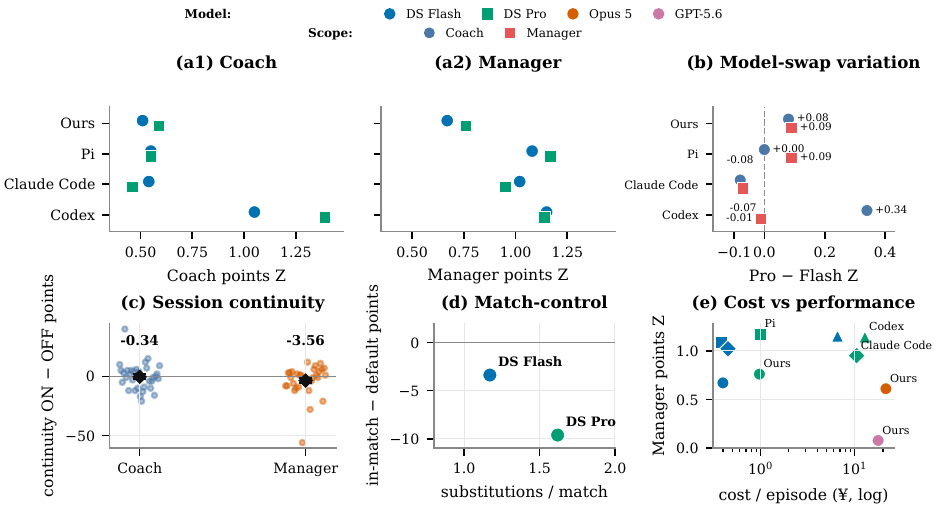}
\caption{\textbf{Scaffold and model contrasts within the Flash--Pro pair.}
Absolute scores and model swaps (a--b) precede continuity, match-control, and
cost diagnostics (c--e); Manager scaffold spans are 0.48/0.41~Z versus a
maximum Manager model swap of 0.09~Z.}
\label{fig:harnessmap}
\end{figure*}

\subsection{Attribution: scaffold and model contrasts}
\label{sec:rq2}

The Agent-Track rows of Table~\ref{tab:main} give the scaffold-by-model matrix.
Pi, Claude Code, and Codex each obtain higher mean Manager scores than Ours,
the fixed stateless scaffold, for both tested models. Paired over the same scenarios and seeds,
the scaffold gains over Ours are $+0.41\pm0.15$, $+0.35\pm0.13$ and
$+0.48\pm0.17$ Z for Flash (Pi, Claude Code, Codex) and $+0.40\pm0.25$,
$+0.18\pm0.22$ and $+0.38\pm0.18$ Z for Pro (mean $\pm$ SE, clustered by seed
as in Table~\ref{tab:leaderboard_uncertainty}). The four Pro-minus-Flash
contrasts range from $-0.08$ to $+0.09$ Z, with seed-clustered SEs of
$0.11$--$0.22$ Z. Holding the model fixed, the Manager scores span $0.48$ Z
for Flash and $0.41$ Z for Pro; the best Coach configuration overall is Codex
($+1.39$). The individual contrasts appear in Table~\ref{tab:paired_contrasts}.

A contemporaneous paired ablation on Flash compares the Claude Code scaffold with
and without session continuity. Coach scores are
effectively unchanged ($+0.60$ versus $+0.61$ Z), while Manager scores show a
directional decrease with continuity ($+0.81$ versus $+1.08$ Z; paired difference
$-3.56\pm2.28$ points). The long-horizon comparison
in Section~\ref{sec:rq3} is a separate experiment whose Claude Code arms use a
no-memory configuration.

\textbf{Finding 3.} \emph{Within the Flash--Pro pair crossed across all
scaffolds, estimated Manager gains over Ours, the stateless scaffold, range from
$0.18$ to $0.48$ Z, while Pro-minus-Flash estimates range from $-0.08$ to
$+0.09$ Z. These contrasts describe the tested configurations. In a
contemporaneous Flash ablation, session continuity leaves Coach performance
unchanged and is directionally worse on Manager.}

\subsection{Supporting diagnostic: match-control granularity}

To test the finest timescale, the same coach episodes are re-run with in-match control enabled: checkpoints at $30'$, half-time, $60'$, $75'$, with
substitutions and live tactic changes, on the same seeds and world. Agents use
the axis: DS Pro substitutes 38\% more often than DS Flash
(1.62 versus 1.17 substitutions per match), but it does not help
them (Figure~\ref{fig:harnessmap}(d)). Across the eight paired episodes, the
control-minus-default difference is $-3.38\pm3.03$ points for DS Flash
(mean$\pm$SE; four negative, three positive, and one tied) and
$-9.63\pm3.75$ for DS Pro (seven negative and one positive).

\textbf{Finding 4.} \emph{Minute-level intervention is actively used but
yields no detectable benefit in season-level outcomes under our current
evaluation: the paired estimate is $-3.38$ points for DS Flash and
$-9.63$ points for DS Pro at $n=8$ pairs each.}

\begin{figure*}[t]
\centering
\includegraphics[width=\textwidth]{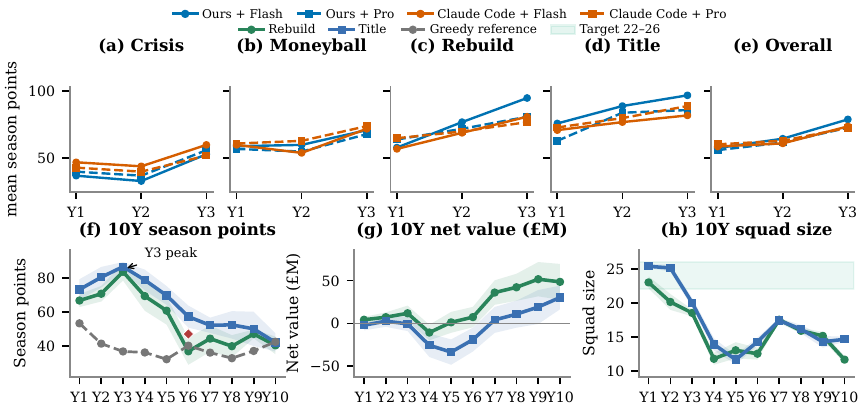}
\caption{\textbf{Early rankings can change over longer horizons.}
Scenario and overall 3Y means (a--e) accompany 10Y sporting, net-value, and
squad-size trajectories (f--h); the selected agent peaks in year~3 and stays below that level in every later year. Bands in (f--h) show $\pm1$ SE
within each scenario; the red diamond marks the cross-scenario mean at year~6.}
\label{fig:horizon}
\end{figure*}

\subsection{RQ2: Temporal persistence across horizons}
\label{sec:rq3}

We ran four scaffold-model combinations over three consecutive seasons (3Y)
across four scenarios. Every configuration contains eight seeds per scenario
(32 trajectories). For the 10Y case study we use Claude Code+Pro, preselected under
the no-memory Manager configuration before the 10Y evaluation. We chose rebuild and title as contrasting management
objectives and ran eight seeds per scenario against the greedy reference. The Claude Code long-horizon episodes use a no-memory configuration rather than
the session-continuity protocol of the main Agent Track. Three-year results change the ranking within the same 3Y cohort
(Figure~\ref{fig:horizon}): Claude Code+Pro leads in year one by
$2.8\pm2.5$ points over Ours+Flash, whereas by year three Ours+Flash finishes
first (79.0, versus 73.8 for Claude Code+Flash, 72.9 for Ours+Pro, and 72.8 for
Claude Code+Pro), leading by $5.3\pm1.9$, $6.2\pm2.8$ and $6.2\pm2.1$ points.
Paired contrasts use 32 matched scenario--seed observations, with standard
errors clustered by seed; across the three seasons the margin in favour of
Ours+Flash widens by $4.6$ to $9.0$ points
(Table~\ref{tab:paired_contrasts}). Within Flash, the scaffold ordering
reverses; within Pro, the two scaffolds converge to effectively the same Y3
score. The ten-year Claude Code curve peaks in year three
(85.4), before its season-points trajectory falls to 47.1 by year six and
remains far below its year-three peak thereafter. This decline coincides with
substantial squad attrition (Appendix~\ref{app:dynasty_finance}). Across ten
seasons the agent accumulates $602.4$ points versus $388.7$ for the greedy
reference, so the decline is relative to its own early peak, not a cumulative
deficit; the reference follows fixed conservative rules and shows no comparable
early peak.

\textbf{Finding 5.} \emph{The 3Y cohort shows a directional reversal in mean
ranking between years one and three. The 10Y case reveals a later decline in
season points, motivating evaluation across multiple horizons.}

\subsection{Supporting diagnostics: cost and resource use}

Costs are converted to a common undiscounted list-price baseline, with
DeepSeek frozen at its pre-August-17 prices. For the reported Model-Track episodes (64 per model), Opus~5 and GPT-5.6 cost approximately 22$\times$ and
18$\times$ as much as DS~Pro, respectively, while scoring below
it on the Manager scope (Figure~\ref{fig:harnessmap}(e)). Across the six stateful Flash/Pro scaffold configurations, cached-input volume
is over two orders of magnitude larger than for the two stateless configurations. The separate no-memory Claude Code
configuration loses to the stateless scaffold at three years.

\textbf{Finding 6.} \emph{For the tested model versions and price snapshot,
price is poorly aligned with Manager performance: DS Pro outperforms Opus~5 and GPT-5.6 while costing 18--22$\times$
less.}

\section{Discussion}

\paragraph{Functional composition.} FromPitch2Board measures whether competence
on an existing responsibility set is preserved as responsibility expands.
GPT-5.6 coaches and recruits effectively at the Recruiter rung, but its
sporting performance and participation decline after expansion to full
management: responsibility expansion coincides with sharply reduced
participation rather than malformed or invalid actions. This passivity extends to shared duties: GPT-5.6's
matchday skip rate rises from 0.2\% as Recruiter to 58.0\% as Manager, rather
than being confined to newly introduced management events
(Appendix~\ref{app:behavior_audit}). Separately, the Flash--Pro model--scaffold
cross shows that agent outcomes can depend materially on scaffold choice,
motivating configuration-level rather than model-only attribution. Functional
composition is therefore an empirical property of an agent configuration, not
a consequence that can be inferred from isolated task scores.

\paragraph{Temporal persistence and scope.} Multi-season results pose the
parallel temporal question: competence observed in year one need not determine
later rankings or trajectories. FM-Bench~\citep{fmbench}, developed concurrently
in a different simulator and interface, reports a related execution gap in
which recorded plans are not subsequently carried out. This supports measuring sustained engagement.
The paired Flash ablation finds session continuity
neutral on Coach and directionally worse on Manager.
The long-horizon evaluation uses a separate
no-memory protocol.

\paragraph{Scope of the measurements.} All results are operational measurements
inside a simulator whose worlds, players, and clubs are procedurally generated
and anonymized. They characterize the tested model versions, scaffolds, and
protocol rather than the providers that serve them, and they are neither
psychological claims about a model nor evidence that any evaluated system is
suitable for real organizational, employment, sporting, or financial decisions.
Running the benchmark consumes computation and paid inference, which is why we
report token and cost accounting alongside the performance results.

\section{Conclusion}
\label{sec:conclusion}

FromPitch2Board uses football management as a controlled testbed for two
questions in long-horizon agency: whether competence is preserved as
heterogeneous responsibilities expand and whether it persists across repeated,
state-dependent decisions. Functional composition is not monotonic: GPT-5.6
improves from Coach to Recruiter, then loses that gain under full management as
its behavior becomes markedly passive. Three-year results show a directional
reversal in mean ranking, and a selected ten-year trajectory peaks early and remains
below that peak. Model--scaffold crosses allow some effects to be localized to a tested factor
rather than treating every agent outcome as a property of the model alone.

\textbf{Limitations and Future Work.} Results cover one simulated management domain, four foundation models, and the
configurations of Section~\ref{sec:experiments}, so cross-domain transfer remains
untested. The model--scaffold cross is complete only for Flash--Pro, and the
ten-year trajectory follows a single scaffold--model combination. Single-season
outcomes remain noisy at eight evaluation seeds per cell. Future work should test engagement-aware scaffolds,
long-horizon memory designs, and whether these patterns generalize across domains
and horizons.

\clearpage
\bibliography{main}

\appendix
\clearpage
\section{Appendix}

\subsection{Benchmark comparison}
\label{app:benchmark_comparison}

\begin{table*}[!tbp]
\centering
\small
\setlength{\tabcolsep}{4pt}
\resizebox{\textwidth}{!}{%
\begin{tabular}{lccccccc}
\toprule
Benchmark & Horizon & Timescale & Scope axis & Harness$\times$model & Det. & Multi-obj. & Same interface\\
\midrule
\textbf{FromPitch2Board} & 1--10y & min$\to$career & \textbf{Coach$\to$Manager} & \textbf{Flash--Pro} & yes & yes & yes\\
FM-Bench~\citep{fmbench} & 5--20y & season$\to$career & no & no & yes & yes & yes\\
Pok\'eAgent~\citep{pokeagent} & long & minute$\to$run & no & partial & no & no & yes\\
CEO/YC~\citep{ceo,yc} & 100s turns & day$\to$year & no & no & NR & partial & yes\\
Enterprise~\citep{enterprise} & 132mo & month$\to$decade & no & no & NR & partial & yes\\
WebArena/OSWorld~\citep{webarena,osworld} & bounded & minutes & no & no & no & no & yes\\
$\tau$-bench~\citep{tau} & session & minutes & no & no & no & no & yes\\
GRF~\citep{grf} & match & minute$\to$match & no & no & NR & no & yes\\
\bottomrule
\end{tabular}%
}
\caption{Comparison with representative agent benchmarks. ``NR'' denotes a dimension not reported by the source paper.}
\label{tab:compare}
\end{table*}

Table~\ref{tab:compare} emphasizes the closest comparisons rather than an
exhaustive leaderboard. The broader lineage includes text-based and open-ended
game environments~\citep{textworld,nethack,crafter,smartplay,gamebench,orak},
updated conversational evaluation~\citep{tau2}, and economic or competitive
multi-agent simulations~\citep{economist,multiagent,cattle}. These works inform
the environment and interaction setting, while our comparison focuses on the
controlled variation of responsibility, scaffold, and horizon.

\subsection{Task interface}
\label{app:interface}

At each decision point the runner provides one JSON observation and accepts one
JSON action. Decisions are triggered on matchdays, upon incoming offers
(\textbf{Manager} only), and on scheduled market days approximately every three
days during open transfer windows (\textbf{Recruiter} and \textbf{Manager}
only). When triggers coincide
the runner ranks them matchday first, offer second, market day third, so one
decision always takes one action. The observation carries the date, step index,
a matchday flag,
league position and points, budget and window state, the squad (opaque id,
name, position, overall rating, age, condition, fitness, morale, injury and
transfer-list flags, wage, market value), the current market listing, any
incoming offers, and the result of the previous action. Market ratings are
hidden until the player is scouted.

\begin{table}[t]
\centering
\small
\caption{Action availability by responsibility scope. The scope is fixed for the
whole episode and enforced by the runner: an out-of-scope action is rejected
with an explanatory message and consumes the decision point. Substitute and
MatchTactics are available to all three responsibility scopes only when
in-match control is enabled.}
\label{tab:permissions}
\begin{tabular}{lccc}
\toprule
Action & \textbf{Coach} & \textbf{Recruiter} & \textbf{Manager} \\
\midrule
Continue (skipped decision) & $\bullet$ & $\bullet$ & $\bullet$ \\
SetLineup              & $\bullet$ & $\bullet$ & $\bullet$ \\
SetTactics             & $\bullet$ & $\bullet$ & $\bullet$ \\
SetMatchPlan           & $\bullet$ & $\bullet$ & $\bullet$ \\
Substitute             & $\bullet$ & $\bullet$ & $\bullet$ \\
MatchTactics           & $\bullet$ & $\bullet$ & $\bullet$ \\
Scout                  & --        & $\bullet$ & $\bullet$ \\
MakeBid                & --        & $\bullet$ & $\bullet$ \\
AcceptOffer            & --        & --        & $\bullet$ \\
RejectOffer            & --        & --        & $\bullet$ \\
CounterOffer           & --        & --        & $\bullet$ \\
ListPlayer             & --        & --        & $\bullet$ \\
\bottomrule
\end{tabular}
\end{table}

Within a match the runner stops the clock only for Substitute, MatchTactics and
Continue, the three in-match actions that every scope holds.

\begin{verbatim}
observation  {"step": 0, "date": "2026-07-01",
              "is_matchday": true, "points": 0,
              "league_position": 19, "budget": 5000000,
              "transfer_window_open": true, "offers": [],
              "next_fixture":
                "2026-07-01 Club_06 vs Club_32 (A)",
              "squad": [{"id": "bc351785-...",
                         "name": "Player_683",
                         "position": "Goalkeeper", "ovr": 73,
                         "age": 34, "condition": 89,
                         "fitness": 75, "morale": 48,
                         "injured": false,
                         "transfer_listed": false,
                         "wage": 5329,
                         "market_value": 1065800}, ...]}
action       {"action": "SetMatchPlan",
              "params": {"player_ids":
                ["bc351785-...", ...11 ids...],
                "play_style": "Defensive"}}
environment  "Set lineup (11 of 11 provided ids are in
              your squad) and Defensive tactics for the
              next match."
\end{verbatim}

\paragraph{Objective.} Each scenario ships a qualitative goal in the prompt
(e.g.\ \emph{avoid relegation} in the crisis scenario, \emph{finish in the top
half while keeping net transfer spend small} in the moneyball scenario), and
episodes are scored by league points. The agent never sees the calibration
statistics or the normalisation that builds $Z$.

\paragraph{Runner rules.} The action set contains no contract renewal and the
observation does not expose contract end dates: a player whose contract expires
leaves the squad automatically, and the agent sees only that the player is gone.
Squad maintenance therefore runs through bids, sales and transfer-listing, never
through retention. Lineups are slot-aligned: the engine keeps the requested
starters and fills every empty slot with the best available player for that
slot, and if fewer than eight of the requested players are still available it
discards the request and selects an automatic best-fit eleven. No
minimum-squad rule applies and no error is raised, so a squad that cannot field
eleven players degrades silently.

\paragraph{Greedy reference.} The greedy reference is a scripted policy that
sees exactly the same observation as the LLM agents and consumes one action per
decision point. At each decision point it enumerates the actions its scope
permits, scores the resulting selections with a hand-written utility
$U = \mathrm{ovr} + \mathrm{role\_fit} - 0.5(100-\mathrm{condition})
- 0.2(100-\mathrm{fitness})$, where $\mathrm{role\_fit}$ subtracts $8$ for
fielding a player out of position and injured players are excluded, and takes
the highest-scoring action. In the \textbf{Manager} scope it additionally treats
a position group as needy while that group's best player is rated below the
squad average plus four, scouts before bidding, prices bids for value, accepts
offers at $1.1\times$ valuation for surplus players and $1.8\times$ for
starters, and lists non-starters aged $27$ or over.

\subsection{Experiment matrix}
\label{app:matrix}

Table~\ref{tab:matrix} lists the axis each experiment varies and the
configurations it is run on. Only the Flash--Pro model pair is crossed with all
four scaffolds. Every other axis is varied one at a time, so the design
attributes differences to a tested factor within the stated configurations
rather than decomposing variance over the full product of axes.

\begin{table}[!tbp]
\centering
\small
\setlength{\tabcolsep}{4pt}
\begin{tabular}{@{}p{2.35cm}p{2.35cm}p{3.1cm}@{}}
\toprule
Experiment & Factor varied (levels) & Run on\\
\midrule
Main leaderboard, Model Track & foundation model (4) & Ours scaffold\\
Main leaderboard, Agent Track & scaffold (4) & Flash--Pro pair\\
Responsibility ladder & responsibility scope (3) & DS Pro stateless; Pi; GPT-5.6\\
Match-control granularity & in-match control (2) & DS Flash; DS Pro\\
Session continuity & session memory (2) & Claude Code + Flash\\
Thinking budget & reasoning cap (3) & DS Flash; DS Pro\\
Horizon & seasons (1/3/10) & 3Y: four configurations; 10Y: one\\
\bottomrule
\end{tabular}
\caption{Experiment matrix. The leaderboard crosses four scaffolds with the
Flash--Pro model pair; the remaining axes are varied on the configurations
listed.}
\label{tab:matrix}
\end{table}

\subsection{Per-scenario Model-Track results}
\label{app:model_scenarios}

Table~\ref{tab:model_scenarios} reports points Z for each scenario. Each entry
uses eight evaluation seeds; Table~\ref{tab:main} is the equal-weight mean of
the four unrounded scenario statistics in the corresponding row. Scenario
entries are rounded independently for display and therefore need not reproduce
the aggregate rounding exactly.

\begin{table}[!tbp]
\centering
\small
\setlength{\tabcolsep}{3pt}
\begin{tabular}{lcccc}
\toprule
Model / scope & crisis & moneyball & rebuild & title\\
\midrule
DS Flash / Coach   & $+0.49$ & $+0.61$ & $+0.79$ & $+0.16$\\
DS Flash / Manager & $+0.19$ & $+0.27$ & $+1.28$ & $+0.95$\\
DS Pro / Coach     & $+0.30$ & $+0.62$ & $+0.95$ & $+0.48$\\
DS Pro / Manager   & $+0.57$ & $+0.97$ & $+0.64$ & $+0.87$\\
Opus 5 / Coach     & $+0.38$ & $+0.85$ & $+0.77$ & $+0.19$\\
Opus 5 / Manager   & $+0.30$ & $+0.25$ & $+0.88$ & $+1.03$\\
GPT-5.6 / Coach    & $+0.25$ & $+1.17$ & $+0.99$ & $+0.41$\\
GPT-5.6 / Manager  & $-0.20$ & $-0.01$ & $+0.10$ & $+0.43$\\
\bottomrule
\end{tabular}
\caption{Per-scenario Model-Track points Z-scores (8 seeds per entry).}
\label{tab:model_scenarios}
\end{table}

\paragraph{Leaderboard uncertainty and scale sensitivity.}
Table~\ref{tab:leaderboard_uncertainty} reports standard errors obtained by
treating each evaluation seed as a cluster containing its four
scenario-normalized scores. The frozen calibration statistics are held fixed.
This uncertainty describes variation over the eight evaluation seeds, rather
than uncertainty in the 100-seed calibration distribution.

\begin{table}[!tbp]
\centering
\small
\setlength{\tabcolsep}{4pt}
\begin{tabular}{llrr}
\toprule
Scaffold & Model & Coach Z $\pm$ SE & Manager Z $\pm$ SE\\
\midrule
Ours & DS Flash & $0.510\pm0.158$ & $0.673\pm0.187$\\
Ours & DS Pro & $0.589\pm0.157$ & $0.764\pm0.240$\\
Ours & Opus 5 & $0.547\pm0.188$ & $0.615\pm0.142$\\
Ours & GPT-5.6 & $0.704\pm0.158$ & $0.080\pm0.155$\\
Pi & DS Flash & $0.554\pm0.126$ & $1.079\pm0.125$\\
Pi & DS Pro & $0.552\pm0.207$ & $1.165\pm0.109$\\
Claude Code & DS Flash & $0.543\pm0.132$ & $1.022\pm0.177$\\
Claude Code & DS Pro & $0.462\pm0.173$ & $0.945\pm0.253$\\
Codex & DS Flash & $1.047\pm0.133$ & $1.152\pm0.167$\\
Codex & DS Pro & $1.392\pm0.225$ & $1.143\pm0.145$\\
\bottomrule
\end{tabular}
\caption{Seed-clustered uncertainty for the main points leaderboard. Each
estimate uses four scenarios and eight seeds per scenario.}
\label{tab:leaderboard_uncertainty}
\end{table}

As a scale sensitivity check, the Model-Track raw Coach/Manager means are
43.56/54.94 (DS Flash), 44.69/55.72 (DS Pro), 44.06/53.88 (Opus~5), and
46.12/46.81 (GPT-5.6). GPT-5.6 gains only $0.69\pm1.77$ raw points from Coach to Manager, whereas the
other three models gain 9.81--11.38. The positive normalized Composition Gap
therefore denotes lost advantage relative to the Manager reference, while the
separate Recruiter comparison localizes the loss of the recruitment gain.

\subsection{Session-continuity ablation}
\label{app:continuity}

We evaluate the Flash session scaffold with and without continuity on the same
scenarios and eight seeds. The two arms are identical except for whether the scaffold carries its
session context across turns. Coach changes by only $-0.34$ points on average,
whereas Manager changes by $-3.56$ points (paired standard error $2.28$), or
approximately $-0.27$ Z after scenario-specific normalization. This supports a directional Manager-specific effect.

\subsection{Thinking-budget ablation}

A thinking-budget ablation (2048-token cap, 8192-token cap, and non-binding) on
title+rebuild $\times$ 8 seeds shows sensitivity to the cap. DS Pro gains 8.2
points from 2048 to 8192, while DS Flash spans 5.4 points across the three
settings. This sensitivity motivates using an aligned, non-binding reasoning
protocol in the main comparison.

\subsection{Token, cost, and runtime accounting}
\label{app:cost_accounting}

Table~\ref{tab:cost_accounting} reports token, runtime, and cost accounting by configuration. One episode spans one configuration, one scenario, and one seed. Dashes denote inapplicable scores. DeepSeek
costs use the pre-August-17 list prices (Pro:
\textyen3/\textyen6; Flash: \textyen1/\textyen2 per million input/output
tokens, with the recorded cache-read price); no paid-bill discount is applied.
Opus~5 uses its official \$5/\$25 input/output list price. GPT-5.6 uses its
original \$5/\$0.50/\$30 uncached-input/cached-input/output list price rather
than the later promotional price. The gateway that served Opus~5 and GPT-5.6
billed these two models at 0.29 of the official list prices at this snapshot,
and it may omit billed reasoning tokens from its returned usage objects, so we
recover both costs as the measured platform bill divided by 0.29. The GPT-5.6 Recruiter evaluation maps its
\textyen300 invoice to \textyen1,034 on the same frozen official-price
baseline. All rows consequently represent undiscounted list-price-equivalent
costs at the declared price snapshot.

\begin{table*}[!tbp]
\centering
\scriptsize
\setlength{\tabcolsep}{3pt}
\resizebox{\textwidth}{!}{%
\begin{tabular}{lrrrrrrrr}
\toprule
Configuration & Episodes & Manager Z & Input & Output & Cache read & Wall time & Cost (\textyen)\\
\midrule
Ours + DS Flash & 64 & $+0.67$ & 11.8M & 7.08M & 1.67M & 1,058s & 25.5\\
Ours + DS Pro & 64 & $+0.76$ & 11.4M & 4.88M & 1.60M & 1,017s & 62.3\\
Claude Code + DS Flash (session continuity) & 64 & $+1.02$ & 15.8M & 1.83M & 467.6M & 322s & 28.9\\
Claude Code + DS Pro (session continuity) & 64 & $+0.95$ & 215.6M & 2.34M & 434.2M & 661s & 671.8\\
Pi + DS Flash & 64 & $+1.08$ & 13.8M & 1.62M & 374.1M & 397s & 24.5\\
Pi + DS Pro & 64 & $+1.17$ & 13.8M & 2.11M & 386.2M & 532s & 63.8\\
Pi + DS Pro (Recruiter rung) & 32 & --- & 10.4M & 0.92M & 278.2M & 532s & 43.8\\
Claude Code + DS Flash (continuity on, paired rerun) & 64 & $+0.81$ & 15.4M & 2.09M & 439.3M & 340s & 28.4\\
Claude Code + DS Flash (continuity off, paired rerun) & 64 & $+1.08$ & 15.6M & 2.06M & 451.7M & 334s & 28.9\\
Codex + DS Flash        & 64 & $+1.15$ & 408.9M & 2.43M & 363.9M & 830s & 421.0\\
Codex + DS Pro          & 64 & $+1.14$ & 266.1M & 3.08M & 243.9M & 1,107s & 822.8\\
Opus 5                  & 64 & $+0.61$ & 35.2M & 0.62M & 0.01M & 508s & 1,375\\
GPT-5.6                 & 64 & $+0.08$ & 32.2M & 1.36M & 10.7M & 730s & 1,138\\
GPT-5.6 (Recruiter rung) & 32 & --- & 9.29M & 0.52M & 0.13M & 588s & 1,034\\
\bottomrule
\end{tabular}%
}
\caption{Token, runtime, and normalized-cost accounting. Token and cost columns
aggregate the episodes listed for each configuration; wall time is the mean per
episode rather than a batch elapsed time. Dashes mark inapplicable scores.}
\label{tab:cost_accounting}
\end{table*}

Additional aggregate-only cost records are \textyen470.5 for the four 3Y configurations, \textyen352.0 for the 16-episode Claude Code+DS Pro 10Y evaluation, and
\textyen136.4 for the ladder and 8192-token-budget episodes.

The DeepSeek rows in Table~\ref{tab:cost_accounting} sum to
\textyen2,221.7, and the three additional aggregate records above sum to
\textyen958.9. Adding \textyen1,375 for Opus~5 and \textyen2,172 for the two
GPT-5.6 evaluations gives \textyen7,413 overall. The table separates measured token counts from billed-cost normalization: cache reads can be large for session
scaffolds without contributing proportionally to cost because their unit
price is much lower than fresh input.

\FloatBarrier
\subsection{Environment baselines}
\label{app:validity}

We first check that the environment orders weak and strong policies as expected. Table~\ref{tab:env_baselines}
shows a monotone capability ladder on both scopes. These are simulation controls and incur no inference cost.

\begin{table}[!tbp]
\centering
\small
\setlength{\tabcolsep}{6pt}
\begin{tabular}{llr}
\toprule
Scope & Policy & Points Z\tabularnewline
\midrule
Coach & Random actions & $-1.83$\tabularnewline
Coach & Best-XI heuristic & $-0.30$\tabularnewline
Coach & Greedy reference & $+0.08$\tabularnewline
\midrule
Manager & Random manager & $-2.83$\tabularnewline
Manager & Passive / no-op & $-0.51$\tabularnewline
Manager & Greedy reference & $-0.09$\tabularnewline
\bottomrule
\end{tabular}
\caption{Environment baseline ordering. The expected weak-to-strong ordering
holds on both responsibility scopes.}
\label{tab:env_baselines}
\end{table}
\subsection{Full responsibility-ladder diagnostics}
\label{app:ladder}

The DS Pro stateless ladder is run as a separate cohort with eight seeds per
scenario (96 episodes). For Pi and GPT-5.6, the Recruiter rung is evaluated on
the same scenarios and seeds as the Coach and Manager endpoints reported in the
main tracks (Tables~\ref{tab:ladder_pi} and~\ref{tab:ladder_gpt}).

\begin{table}[!tbp]
\centering
\small
\setlength{\tabcolsep}{4pt}
\begin{tabular}{lrrrrrr}
\toprule
Responsibility & Overall & Crisis & Moneyball & Rebuild & Title & Steps\tabularnewline
\midrule
Coach only & 45.0 & 35 & 50 & 50 & 44 & 38.0\tabularnewline
+ Recruitment & 58.4 & 41 & 65 & 57 & 71 & 47.0\tabularnewline
+ Full management & 56.8 & 38 & 65 & 55 & 69 & 47.9\tabularnewline
\bottomrule
\end{tabular}
\caption{DS Pro stateless responsibility ladder by scenario (eight seeds each;
96 episodes). Overall and scenario means are rounded independently.}
\label{tab:ladder_full}
\end{table}

\begin{table}[!tbp]
\centering
\small
\setlength{\tabcolsep}{4pt}
\begin{tabular}{lrrrrr}
\toprule
Responsibility & Overall & Crisis & Moneyball & Rebuild & Title\tabularnewline
\midrule
Coach only & 44.1 & 40.6 & 46.4 & 42.9 & 46.4\tabularnewline
+ Recruitment & 61.6 & 37.4 & 65.9 & 66.0 & 77.0\tabularnewline
+ Full management & 60.9 & 40.8 & 67.4 & 60.9 & 74.6\tabularnewline
\bottomrule
\end{tabular}
\caption{DS Pro responsibility ladder with Pi (eight seeds per scenario),
where Coach and Manager correspond to the Agent-Track evaluations.}
\label{tab:ladder_pi}
\end{table}

\begin{table}[!tbp]
\centering
\small
\setlength{\tabcolsep}{4pt}
\begin{tabular}{lrrrrr}
\toprule
Responsibility & Overall & Crisis & Moneyball & Rebuild & Title\tabularnewline
\midrule
Coach only & 46.1 & 36.9 & 53.3 & 47.5 & 46.9\tabularnewline
+ Recruitment & 58.1 & 41.9 & 55.1 & 62.8 & 72.6\tabularnewline
+ Full management & 46.8 & 32.0 & 49.3 & 48.8 & 57.3\tabularnewline
\bottomrule
\end{tabular}
\caption{GPT-5.6 responsibility ladder (eight seeds per scenario). Recruiter
exceeds Manager by $11.28 \pm 2.28$ paired points ($n=32$).}
\label{tab:ladder_gpt}
\end{table}

\begin{table*}[!tbp]
\centering
\small
\setlength{\tabcolsep}{7pt}
\begin{tabular}{lrrrr}
\toprule
Outcome & Recruiter & Manager & Manager $-$ Recruiter & Paired SE\tabularnewline
\midrule
Points & 58.09 & 46.81 & $-11.28$ & 2.28\tabularnewline
League position $\downarrow$ & 9.13 & 13.31 & $+4.19$ & 1.00\tabularnewline
Goal difference & 8.75 & $-13.00$ & $-21.75$ & 4.05\tabularnewline
Balance (\pounds M) & 61.53 & 67.32 & $+5.79$ & 2.08\tabularnewline
Net value (\pounds M) & 9.66 & 8.83 & $-0.83$ & 2.07\tabularnewline
Net spend (\pounds M) & 10.43 & 4.70 & $-5.73$ & 2.08\tabularnewline
Wage bill (\pounds k) & 329.47 & 296.41 & $-33.06$ & 5.24\tabularnewline
Squad value (\pounds M) & 65.89 & 59.28 & $-6.61$ & 1.05\tabularnewline
Average age & 24.83 & 25.04 & $+0.20$ & 0.07\tabularnewline
Squad size & 23.53 & 22.03 & $-1.50$ & 0.21\tabularnewline
\bottomrule
\end{tabular}
\caption{GPT-5.6 Recruiter-to-Manager paired outcomes ($n=32$ matched episodes). Full
management improves liquidity but reduces sporting and squad-value outcomes.}
\label{tab:gpt_multiobjective}
\end{table*}

For the stateless DS Pro ladder, the final responsibility rung changes more
than points. Relative to recruitment
only, net spend falls from \pounds17.1M to \pounds8.2M, net value rises from
\pounds2.9M to \pounds9.7M, squad value falls from \pounds65.7M to
\pounds63.1M, and squad size contracts from 23.5 to 22.3 players. Thus, the
$-1.6$-point change accompanies a measurable financial and roster trade-off,
rather than an across-the-board regression.

\FloatBarrier
\subsection{Protocol-alignment experiments}
\label{app:protocol}

\paragraph{Main evaluation configuration.} The served DeepSeek versions are
\texttt{DeepSeek\allowbreak-V4\allowbreak-Flash\allowbreak-0731} and \texttt{DeepSeek\allowbreak-V4\allowbreak-Pro\allowbreak-0813}, abbreviated
as DS Flash and DS Pro. The other served identifiers are \texttt{claude\allowbreak-opus\allowbreak-5} and \texttt{gpt\allowbreak-5.6\allowbreak-sol}, abbreviated as Opus~5 and GPT-5.6. All
four models use the same stateless scaffold and declared maximum-reasoning
setting, with non-binding token ceilings. The Agent Track
evaluates Ours, Pi, Claude Code, and Codex through the shared task interface.

\begin{table}[!tbp]
\centering
\small
\setlength{\tabcolsep}{4pt}
\begin{tabular}{llrr}
\toprule
Scaffold/model & Reasoning protocol & Coach Z & Manager Z\tabularnewline
\midrule
Ours / Flash & 2048-token cap & $+0.49$ & $+0.87$\tabularnewline
Ours / Flash & non-binding & $+0.51$ & $+0.67$\tabularnewline
Ours / Pro & 2048-token cap & $+0.44$ & $+0.46$\tabularnewline
Ours / Pro & non-binding & $+0.59$ & $+0.76$\tabularnewline
\midrule
Pi / Flash & minimal & $+0.66$ & $+1.08$\tabularnewline
Pi / Flash & max & $+0.55$ & $+1.08$\tabularnewline
Pi / Pro & minimal & $+0.76$ & $+1.25$\tabularnewline
Pi / Pro & max & $+0.55$ & $+1.17$\tabularnewline
\bottomrule
\end{tabular}
\caption{Protocol-alignment arms motivating the shared non-binding,
max-reasoning protocol used by the main leaderboard.}
\label{tab:protocol_alignment}
\end{table}

\begin{table}[!tbp]
\centering
\small
\begin{tabular}{lrrr}
\toprule
Model & 2048-token cap & 8192-token cap & Non-binding\tabularnewline
\midrule
DS Pro & 56.6 & 64.8 & 59.8\tabularnewline
DS Flash & 66.9 & 61.5 & 65.1\tabularnewline
\bottomrule
\end{tabular}
\caption{Thinking-budget ablation on title and rebuild (eight seeds each),
reported as raw mean points.}
\label{tab:thinking_budget}
\end{table}

\begin{table}[!tbp]
\centering
\small
\setlength{\tabcolsep}{5pt}
\begin{tabular}{llrr}
\toprule
Model & Session continuity & Coach Z & Manager Z\tabularnewline
\midrule
DS Flash & continuity off & $+0.61$ & $+1.08$\tabularnewline
DS Flash & continuity on & $+0.60$ & $+0.81$\tabularnewline
\bottomrule
\end{tabular}
\caption{Contemporaneous Flash session-continuity ablation.}
\label{tab:memory_numeric}
\end{table}

\begin{table*}[!tbp]
\centering
\small
\setlength{\tabcolsep}{4pt}
\begin{tabular}{lrrrrrr}
\toprule
& \multicolumn{3}{c}{DS Flash} & \multicolumn{3}{c}{DS Pro}\\
\cmidrule(lr){2-4}\cmidrule(lr){5-7}
Scenario / seed & Default & Control & $\Delta$ & Default & Control & $\Delta$\tabularnewline
\midrule
Rebuild / 42 & 41 & 41 & 0 & 43 & 28 & $-15$\tabularnewline
Rebuild / 43 & 62 & 49 & $-13$ & 66 & 48 & $-18$\tabularnewline
Rebuild / 44 & 32 & 29 & $-3$ & 31 & 30 & $-1$\tabularnewline
Rebuild / 45 & 45 & 49 & $+4$ & 52 & 50 & $-2$\tabularnewline
Title / 42 & 37 & 41 & $+4$ & 48 & 32 & $-16$\tabularnewline
Title / 43 & 50 & 35 & $-15$ & 48 & 29 & $-19$\tabularnewline
Title / 44 & 38 & 45 & $+7$ & 43 & 53 & $+10$\tabularnewline
Title / 45 & 40 & 29 & $-11$ & 43 & 27 & $-16$\tabularnewline
\midrule
Mean $\pm$ SE & \multicolumn{3}{c}{$-3.38\pm3.03$} & \multicolumn{3}{c}{$-9.63\pm3.75$}\tabularnewline
\bottomrule
\end{tabular}
\caption{Match-control comparison on the eight paired episodes. $\Delta$ is
control minus the default pre-match-only Coach score.}
\label{tab:match_control_pairs}
\end{table*}

\FloatBarrier
\begin{table*}[!tbp]
\centering
\scriptsize
\setlength{\tabcolsep}{3pt}
\begin{tabular}{lcccc}
\toprule
Configuration & Crisis Y1/Y2/Y3 & Moneyball Y1/Y2/Y3 &
Rebuild Y1/Y2/Y3 & Title Y1/Y2/Y3\tabularnewline
\midrule
Greedy & 33/19/20 & 41/32/33 & 53/40/38 & 54/42/35\tabularnewline
Claude Code + Flash (no memory) & 47/44/60 & 60/54/72 & 57/69/81 & 71/77/82\tabularnewline
Claude Code + Pro (no memory) & 43/40/52 & 61/63/74 & 65/70/77 & 73/80/89\tabularnewline
Ours + Flash & 37/33/53 & 59/60/71 & 58/77/95 & 76/89/97\tabularnewline
Ours + Pro & 40/37/56 & 57/55/68 & 64/72/81 & 63/84/86\tabularnewline
\bottomrule
\end{tabular}
\caption{Three-season scenario means (eight seeds per scenario; 32 trajectories
per configuration). Figure~\ref{fig:horizon} uses unrounded values.}
\label{tab:three_year_full}
\end{table*}

\begin{table}[!tbp]
\centering
\small
\setlength{\tabcolsep}{3.5pt}

\begin{tabular}{lrrrrrrrrrr}
\toprule
& Y1 & Y2 & Y3 & Y4 & Y5 & Y6 & Y7 & Y8 & Y9 & Y10 \tabularnewline
\midrule

Claude Code + Pro
& 70.1 & 75.8 & 85.4 & 74.2 & 65.4
& 47.1 & 48.3 & 46.2 & 48.6 & 41.3 \tabularnewline

Greedy
& 53.4 & 41.4 & 36.8 & 36.2 & 32.2
& 40.2 & 36.1 & 32.8 & 37.1 & 42.5 \tabularnewline

\midrule

Claude Code + Pro, cumulative
& 70.1 & 145.9 & 231.3 & 305.5 & 370.9
& 418.0 & 466.3 & 512.5 & 561.1 & 602.4 \tabularnewline

Greedy, cumulative
& 53.4 & 94.8 & 131.6 & 167.8 & 200.0
& 240.2 & 276.3 & 309.1 & 346.2 & 388.7 \tabularnewline

\bottomrule
\end{tabular}

\caption{Ten-season points, averaged over rebuild and title
(eight seeds per scenario in every year). The agent peaks
in year 3 and subsequently declines, finishing slightly
below the greedy reference in year 10; across the full decade it nevertheless
accumulates substantially more points than the reference ($602.4$ against
$388.7$), so the decline is relative to its own peak rather than a cumulative
deficit. Cumulative rows
sum the displayed annual means.}

\label{tab:ten_year_full}
\end{table}

\subsection{Per-scenario Agent-Track results}
\label{app:agent_scenarios}

Table~\ref{tab:agent_scenarios} expands the Agent Track across scenarios. The
scaffold advantage is not confined to one favorable budget condition: all six
external-scaffold configurations exceed their corresponding stateless Manager
aggregate, although the magnitude and the strongest scaffold vary by
scenario.

\begin{table*}[!tbp]
\centering
\small
\setlength{\tabcolsep}{5pt}
\begin{tabular}{llrrrr}
\toprule
Scaffold/model & Scope & Crisis & Moneyball & Rebuild & Title\tabularnewline
\midrule
Pi / Flash & Coach & $+0.38$ & $+0.78$ & $+0.91$ & $+0.14$\tabularnewline
Pi / Flash & Manager & $+0.17$ & $+1.40$ & $+1.31$ & $+1.43$\tabularnewline
Pi / Pro & Coach & $+0.59$ & $+0.61$ & $+0.63$ & $+0.38$\tabularnewline
Pi / Pro & Manager & $+0.58$ & $+1.31$ & $+0.90$ & $+1.87$\tabularnewline
\midrule
Claude Code / Flash & Coach & $+0.36$ & $+1.04$ & $+0.61$ & $+0.16$\tabularnewline
Claude Code / Flash & Manager & $+0.67$ & $+0.96$ & $+0.69$ & $+1.76$\tabularnewline
Claude Code / Pro & Coach & $+0.44$ & $+0.39$ & $+0.71$ & $+0.31$\tabularnewline
Claude Code / Pro & Manager & $+0.35$ & $+0.88$ & $+0.97$ & $+1.59$\tabularnewline
\midrule
Codex / Flash & Coach & $+0.33$ & $+1.32$ & $+1.10$ & $+1.44$\tabularnewline
Codex / Flash & Manager & $+0.76$ & $+0.90$ & $+0.92$ & $+2.02$\tabularnewline
Codex / Pro & Coach & $+1.34$ & $+1.70$ & $+1.61$ & $+0.92$\tabularnewline
Codex / Pro & Manager & $+0.42$ & $+1.21$ & $+1.51$ & $+1.43$\tabularnewline
\bottomrule
\end{tabular}
\caption{Per-scenario Agent-Track points Z-scores, computed against the
frozen responsibility-scope- and scenario-specific calibration (eight seeds per entry).}
\label{tab:agent_scenarios}
\end{table*}

\begin{table}[!tbp]
\centering
\small
\setlength{\tabcolsep}{4pt}
\begin{tabular}{lrrrr}
\toprule
Contrast & $\Delta$ & SE$_{\mathrm{seed}}$ & SE$_{\mathrm{pair}}$ & $t$\\
\midrule
\multicolumn{5}{l}{\emph{Manager points Z} (32 paired cells, 8 seed clusters)}\\
Pi $-$ Ours (Flash) & $+0.41$ & $0.15$ & $0.17$ & $2.7$\\
Claude Code $-$ Ours (Flash) & $+0.35$ & $0.13$ & $0.18$ & $2.7$\\
Codex $-$ Ours (Flash) & $+0.48$ & $0.17$ & $0.22$ & $2.9$\\
Pi $-$ Ours (Pro) & $+0.40$ & $0.25$ & $0.19$ & $1.6$\\
Claude Code $-$ Ours (Pro) & $+0.18$ & $0.22$ & $0.19$ & $0.8$\\
Codex $-$ Ours (Pro) & $+0.38$ & $0.18$ & $0.21$ & $2.1$\\
\midrule
Pro $-$ Flash (Ours) & $+0.09$ & $0.15$ & $0.18$ & $0.6$\\
Pro $-$ Flash (Pi) & $+0.09$ & $0.11$ & $0.18$ & $0.8$\\
Pro $-$ Flash (Claude Code) & $-0.08$ & $0.22$ & $0.15$ & $-0.4$\\
Pro $-$ Flash (Codex) & $-0.01$ & $0.14$ & $0.24$ & $-0.1$\\
\midrule
\multicolumn{5}{l}{\emph{Year-one points, 3Y cohort} (32 paired cells, 8 seed clusters)}\\
Ours+Flash $-$ Claude Code+Flash & $-1.1$ & $2.3$ & $2.4$ & $-0.5$\\
Ours+Flash $-$ Ours+Pro & $+1.6$ & $1.0$ & $2.1$ & $1.5$\\
Ours+Flash $-$ Claude Code+Pro & $-2.8$ & $2.5$ & $2.3$ & $-1.1$\\
\midrule
\multicolumn{5}{l}{\emph{Year-three points, 3Y cohort} (32 paired cells, 8 seed clusters)}\\
Ours+Flash $-$ Claude Code+Flash & $+5.3$ & $1.9$ & $3.3$ & $2.8$\\
Ours+Flash $-$ Ours+Pro & $+6.2$ & $2.8$ & $3.0$ & $2.2$\\
Ours+Flash $-$ Claude Code+Pro & $+6.2$ & $2.1$ & $3.0$ & $3.0$\\
\midrule
\multicolumn{5}{l}{\emph{Year three minus year one, 3Y cohort} (interaction)}\\
Ours+Flash $-$ Claude Code+Flash & $+6.4$ & $4.0$ & $3.9$ & $1.6$\\
Ours+Flash $-$ Ours+Pro & $+4.6$ & $2.7$ & $3.9$ & $1.7$\\
Ours+Flash $-$ Claude Code+Pro & $+9.0$ & $3.7$ & $3.9$ & $2.4$\\
\bottomrule
\end{tabular}
\caption{Paired contrasts.
Each comparison matches configurations by
scenario and seed. $\mathrm{SE}_{\mathrm{seed}}$ clusters
observations by seed, while
$\mathrm{SE}_{\mathrm{pair}}$ treats scenario--seed
differences as independent.
Three-year contrasts use 32 matched observations. The final block reports the
year-three-minus-year-one interaction, that is how much each margin shifts
between the two seasons, so a positive value is a widening lead rather than a
single-season difference.}
\label{tab:paired_contrasts}
\end{table}

\FloatBarrier
\subsection{Event-level behavior analysis}
\label{app:behavior_audit}

The Manager trajectories expose different routes to similar aggregate scores.
Table~\ref{tab:behavior_audit} separates action frequency from execution
success. GPT-5.6's bids are not broadly malformed or ineffective; it enters
the market much less often. DS Pro instead obtains transactions through
repeated participation despite a lower success rate per bid.

\begin{table}[!tbp]
\centering
\small
\setlength{\tabcolsep}{5pt}
\resizebox{\columnwidth}{!}{%
\begin{tabular}{lrrr}
\toprule
Configuration & Skipped decisions & Bid rate / step & Resolved-bid acceptance\tabularnewline
\midrule
Ours + DS Pro / Manager & 16.9\% & 13.9\% & 33\%\tabularnewline
GPT-5.6 / Recruiter & 1.1\% & 12.6\% & 35.0\%\tabularnewline
GPT-5.6 / Manager & 57.9\% & 2.7\% & 38.7\%\tabularnewline
\bottomrule
\end{tabular}
}
\caption{Action profiles. \emph{Skipped decisions} counts decision points at
which the agent issued no action and continued. GPT rates aggregate 32 paired
episodes per responsibility rung; acceptance uses bids with explicit outcomes.}
\label{tab:behavior_audit}
\end{table}
GPT-5.6's inactivity is not confined to the responsibilities that the
Manager rung adds. On the 1,216 matchday decision points that both rungs
contain, its skipped rate rises from 0.2\% to 58.0\%, and market-day
participation falls by a similar factor, so the decline covers the duties it
performed as a Recruiter. Incoming offers, the decision type unique to the
Manager rung, account for only 18 of its 1,544 decision points.

\begin{table}[!tbp]
\centering
\small
\setlength{\tabcolsep}{5pt}
\begin{tabular}{llrrr}
\toprule
Scope & Trigger & Decision points & Skipped & Skipped (\%)\\
\midrule
Recruiter & Matchday & 1,216 & 3 & 0.2\\
Recruiter & Market day & 288 & 14 & 4.9\\
\midrule
Manager & Matchday & 1,216 & 705 & 58.0\\
Manager & Market day & 310 & 179 & 57.7\\
Manager & Incoming offer & 18 & 10 & 55.6\\
\bottomrule
\end{tabular}
\caption{Where GPT-5.6 stops acting, by the event that triggered each decision
point. Triggers are identified by deterministic replay of the recorded action
sequences. Both rungs contain the same 1,216 matchday decision points.}
\label{tab:skip_by_trigger}
\end{table}

\begin{table}[!tbp]
\centering
\small
\setlength{\tabcolsep}{5pt}
\begin{tabular}{lrrrr}
\toprule
Scenario & Recruiter & Manager & $\Delta$ (pp) & Positive pairs\\
\midrule
Crisis & 0.0\% & 57.2\% & $+57.2$ & 8/8\\
Moneyball & 1.0\% & 61.2\% & $+60.2$ & 8/8\\
Rebuild & 0.0\% & 58.6\% & $+58.6$ & 8/8\\
Title & 0.0\% & 54.9\% & $+54.9$ & 8/8\\
\midrule
All 32 pairs & & & median $+57.9$ & 32/32\\
\bottomrule
\end{tabular}
\caption{Matchday skipped-decision rate per episode, paired across the two
responsibility rungs. Every matched episode increases, so the aggregate change
is not driven by a subset of trajectories. The two rungs share the same 1,216
matchday decision points; their state trajectories diverge through the agents'
own decisions.}
\label{tab:passivity_pairs}
\end{table}

\FloatBarrier
\subsection{Sporting--financial divergence over ten seasons}
\label{app:dynasty_finance}

The long-horizon trajectory is multi-objective rather than a monotone exhaustion
of resources. Both scenarios reach their sporting peak in year 3, incur a
sharp contraction in squad size and net value around years 4--6, and later
recover financial net value without recovering league performance
(Table~\ref{tab:dynasty_finance}). This is the delayed-feedback pattern that
is not visible in the one-season score.

\begin{table*}[!tbp]
\centering
\small
\setlength{\tabcolsep}{5pt}
\begin{tabular}{lrrrrrr}
\toprule
Scenario / metric & Y1 & Y3 & Y4 & Y6 & Y9 & Y10\tabularnewline
\midrule
Rebuild points & 66.9 & 83.9 & 69.5 & 36.9 & 47.1 & 40.4\tabularnewline
Rebuild net value (\pounds M) & 4.2 & 11.8 & $-10.7$ & 7.3 & 51.7 & 48.5\tabularnewline
Rebuild squad size & 23.0 & 18.5 & 11.8 & 12.5 & 15.1 & 11.6\tabularnewline
\midrule
Title points & 73.4 & 86.9 & 79.0 & 57.4 & 50.1 & 42.1\tabularnewline
Title net value (\pounds M) & $-2.2$ & $-0.6$ & $-25.6$ & $-19.0$ & 19.1 & 30.2\tabularnewline
Title squad size & 25.4 & 20.0 & 13.9 & 14.2 & 14.2 & 14.6\tabularnewline
\bottomrule
\end{tabular}
\caption{Claude Code+Pro 10Y checkpoints (eight seeds per scenario in every reported year).}
\label{tab:dynasty_finance}
\end{table*}

Contract expiries are the largest recorded outflow during the sharp squad
contraction in years 3--5. Over the first nine
seasons 25.3 players leave on expiry and 21.5 are sold, against 16.3 arrivals;
year 3 alone records 6.9 expiries, 2.4 sales and 1.1 arrivals, and squad size
falls from 22.6 in year 2 to 12.3 in year 5.

\begin{table}[!tbp]
\centering
\small
\setlength{\tabcolsep}{4pt}
\begin{tabular}{lrrrr}
\toprule
Season & Bought & Sold & Released & Squad at end\\
\midrule
1 & 5.6 & 4.1 & 0.0 & 24.2\\
2 & 1.7 & 3.7 & 2.5 & 22.6\\
3 & 1.1 & 2.4 & 6.9 & 19.2\\
4 & 2.7 & 1.7 & 5.6 & 12.8\\
5 & 2.1 & 2.6 & 3.8 & 12.3\\
6 & 1.2 & 1.4 & 0.2 & 13.4\\
7 & 0.6 & 1.2 & 1.3 & 17.4\\
8 & 0.8 & 2.4 & 1.8 & 16.0\\
9 & 0.5 & 2.0 & 3.2 & 14.7\\
\midrule
Total & 16.3 & 21.5 & 25.3 & \\
\bottomrule
\end{tabular}
\caption{Claude Code+Pro roster flow over the first nine seasons, averaged over
rebuild and title (eight seeds per scenario). \emph{Released} denotes contract
expiry. Squad size also moves with youth intake and retirements, which are not
recorded as transfer movements, so the columns need not sum to the change in
squad size.}
\label{tab:roster_flow}
\end{table}

\end{document}